\documentclass[letterpaper, 10 pt, conference]{ieeeconf}  

\IEEEoverridecommandlockouts                              

\usepackage[left=54pt,top=50pt,right=54pt,bottom=57pt]{geometry}
\usepackage{graphicx} 
\usepackage{amsmath} 
\usepackage{amssymb}  
\usepackage{color}
\usepackage[noadjust]{cite}
\usepackage{eqnarray}
\usepackage[ ruled,vlined, noend]{algorithm2e}

\usepackage[colorlinks,bookmarksopen,bookmarksnumbered,citecolor=red,urlcolor=red]{hyperref}

\usepackage{tabularx}
\usepackage{booktabs}
\usepackage[a-1b]{pdfx} 
\usepackage{gensymb}
\usepackage{balance}

\usepackage[colorinlistoftodos,prependcaption,textsize=tiny]{todonotes}

\usepackage[a-1b]{pdfx}

\title{\LARGE \bf
Position Control and Variable-Height Trajectory Tracking of a \\Soft Pneumatic Legged Robot
}

\author{Zhichao Liu and Konstantinos Karydis
\thanks{The authors are with the Dept. of Electrical and Computer Engineering, University of California, Riverside. 
		Email: \{zliu157, karydis\}@ucr.edu. 
	    We gratefully acknowledge the support of NSF \#IIS-1910087 and \#CMMI-2046270, and ARL \#W911NF-18-1-0266.   
		Any opinions, findings, and conclusions or recommendations expressed in this material are those of the authors and do not necessarily reflect the views of the funding agencies.
}}

\begin{document}

\maketitle
\thispagestyle{empty}
\pagestyle{empty}

\begin{abstract}
Soft pneumatic legged robots show promise in their ability to traverse a range of different types of terrain, including natural unstructured terrain met in applications like precision agriculture. They can adapt their body morphology to the intricacies of the terrain at hand, thus enabling robust and resilient locomotion. In this paper we capitalize upon recent developments on soft pneumatic legged robots to introduce a closed-loop trajectory tracking control scheme for operation over flat ground. Closed-loop pneumatic actuation feedback is achieved via a compact and portable pneumatic regulation board. Experimental results reveal that our soft legged robot can precisely control its body height and orientation while in quasi-static operation based on a geometric model. The robot can track both straight line and curved trajectories as well as variable-height trajectories. This work lays the basis to enable autonomous navigation for soft legged robots.
\end{abstract}

\section{Introduction}
Multiple types of rigid robots (e.g., industrial robots used in manufacturing) have been successfully endowed with rapid and precise motion control capabilities~\cite{chiaverini1994review}. However, the high stiffness of the body, as well as the high-gain feedback control can introduce a risk of bodily injuries, especially in cases where interactions with humans are involved~\cite{rus2015design}. In response, in recent years there has been a fast-growing interest in the development and control of \emph{soft robots}. Soft robots can enable safe interaction with humans, high power-to-weight ratio, adaptation to the interacting environment, and comparatively lower fabrication cost~\cite{qiao2019dynamic}. 

Various actuation methods have been developed for soft robots. Some representative examples include pneumatic~\cite{rus2015design, kokkoni2020development}, hydraulic~\cite{marchese2014autonomous}, cable-driven~\cite{bern2019trajectory} and shape-memory alloy (SMA)~\cite{mao2016locomotion} systems. 
Among those methods, pneumatic actuators have been observed to facilitate legged robots' adaptation to various types of terrain, thus making them a suitable candidate for use in the context of robotic locomotion~\cite{drotman20173d, drotman2018application}. Our previous work introduced a soft pneumatic actuator with two degrees of freedom (DoFs) that can both bend and extend to create foot trajectory profiles suitable for legged locomotion~\cite{liu2020sorx}. Utilizing that actuator, we developed a novel soft hexapedal robot (SoRX) that can operate over a range of challenging environments, such as rough, steep, and unstable terrain, without any additional control effort and by following the same feedforward control strategy (an alternating tripod gait scheduler) across all terrains.

However, these soft pneumatic legged robots have limitations as they rely on empirically hand-tuned input sequences for open-loop control. Meanwhile, a lack of mathematical models makes it hard to utilize model-based controllers for precise motion control. Recent related work has introduced a soft pneumatic legged robot powered by electronics-free pneumatic circuits~\cite{drotman2021electronics}. However, the robot still requires tethered manual control for locomotion and collision avoidance.

\begin{figure}[t!]
\vspace{8pt}
	\centering
	\includegraphics[width=0.99\linewidth]{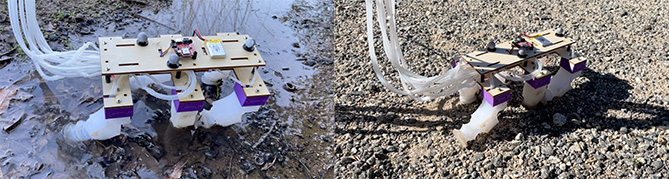}
	\caption{Outdoor operations for SoRX on natural unstructured terrain.}
	\label{fig:cover}
	\vspace{-18pt}
\end{figure}

Model-based motion control for soft pneumatically-actuated robots has been a challenging task due to the nonlinear properties of soft materials and generally slow responses to actuation~\cite{qiao2019dynamic}. Past research on soft pneumatic robots' modeling and control has mostly focused on single actuators or soft manipulators. Model-based control of continuum manipulators with relatively higher stiffness has been well-studied. Piecewise constant curvatures~\cite{ marchese2016design} and variable curvature models~\cite{mahl2014variable} have been proposed to achieve feedforward control. Other attempts include Cosserat rod~\cite{dupont2009design}, mass-damper-spring-based~\cite{guo2015dynamic}, linear parameter-varying~\cite{qiao2019dynamic}, and finite element method-based~\cite{bieze2018finite} models. Those models have then been used to develop various feedforward or feedback control methods, including proportional-integral-derivative (PID)~\cite{bieze2018finite}, sliding mode~\cite{skorina2015feedforward}, model predictive control~\cite{best2016new} and learning-based methods~\cite{thuruthel2018model}.

However, these methods are significantly limited in their application to the control of soft pneumatically-actuated legged robots in three main ways. 1) The methods usually fail to incorporate frequent and periodic interactions with the environment, which are common in legged locomotion. 2) A large majority of methods only take a small number of actuators into account, while controlling soft legged robots is more complicated since the robots usually have at least four legs and each leg has at least two actuated DoFs. 3) The methods require relatively costly and large valves or pressure sources for fast and precise airflow regulation; high cost prohibits scaling to multiple channels of actuation while the size and weight restrict mobility.  


Past research on motion control of soft pneumatic mobile robots has primarily focused on planar locomotion, featuring soft robotic snakes~\cite{onal2013autonomous, qin2018design, luo2020motion}. However, those robots rely on traditional rigid wheels for contacting with the surface, limiting the ability to adapt to various terrain. A recent work presents a continuum soft robot capable of tracking trajectories and interacting with the environment~\cite{della2020model}. Nevertheless, robot movement is still limited to 2D space. 

In this paper, we present a static model for feedforward position control (body height and orientation) of our soft pneumatic legged robot SoRX. In support of our longer-term aim of deploying the robot in outdoor environments, herein we design and develop a low-cost pneumatic regulation board that powers up to eight channels of pressurization/depressurization with air pressure feedback. By utilizing this board, we propose a fast and efficient air pressure feedback controller. Taking advantage of the proposed model and pneumatic regulation system, we further propose a closed-loop trajectory tracking method to enable the robot to track variable-height trajectories trajectories. To the best of our knowledge, SoRX is the first soft pneumatic legged robot to track variable-height trajectories. 

The contributions of this work are as follows: 
\begin{itemize}
	\item We propose a static model based on geometric constraints for feedforward position control (body height and orientation).
	\item We develop a pressure feedback controller based on a custom low-cost pneumatic regulation board with eight channels of pressurization/depressurization.
	\item We introduce a closed-loop trajectory control method to track variable-height trajectories.
	\item We study the robot's position control and trajectory tracking performance experimentally.
\end{itemize}

\section{Modeling and Parameter Identification}\label{sec_model}
\begin{figure}[h!]
\vspace{6pt}
	\centering
	\includegraphics[width=0.75\linewidth]{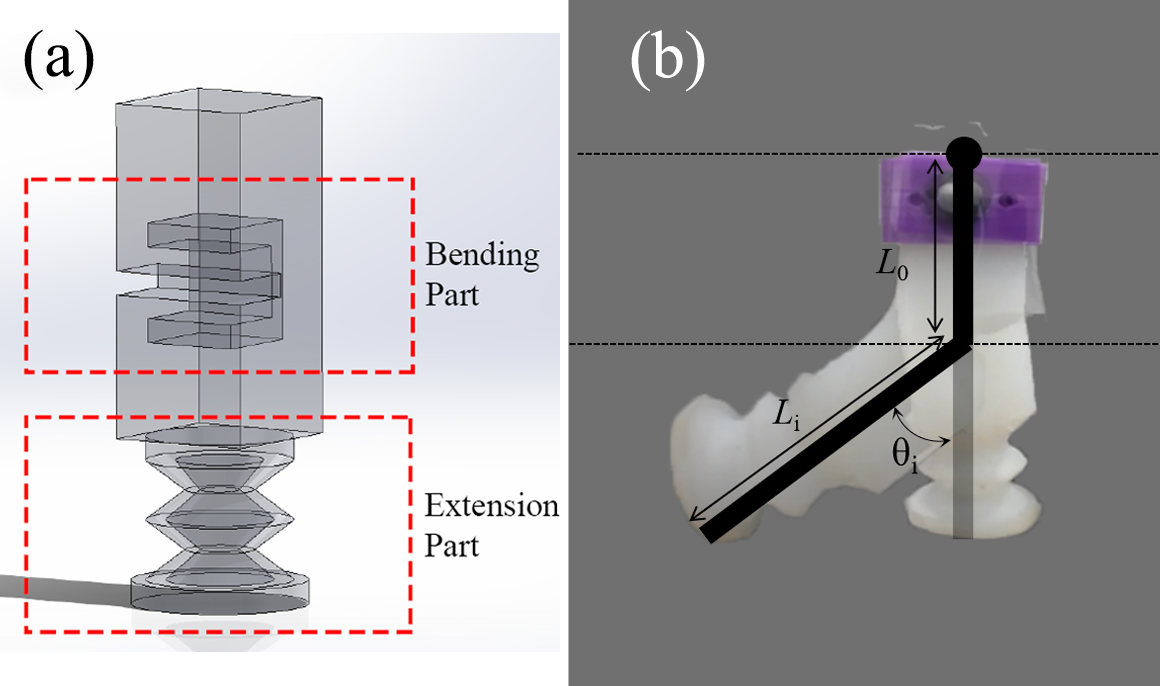}
	\vspace{-6pt}
	\caption{(a) CAD rendering of the leg design, and (b) the proposed static model based on geometric constraints.}
	\label{fig:actuator}
	\vspace{-10pt}
\end{figure}

SoRX has been shown to reach high walking speeds (compared to other soft legged robots) across various types of terrain~\cite{liu2020sorx}. The robot's robust and resilient walking performance mainly comes from the leg design that can bend and extend to create foot trajectory profiles suitable for legged locomotion (see Fig.~\ref{fig:actuator}a). In fact, walking tests in~\cite{liu2020sorx} indicated that SoRX's center of mass (CoM) displays trajectories similar to those observed in traditional hexapedal robots and which are often modeled by the spring-loaded inverted pendulum (SLIP) model~\cite{birkmeyer2009dash}.


However, the SLIP model is unfeasible to be applied on soft pneumatic legged robots for two reasons. First, the weight of legs of SoRX accounts for more than 80\% of the total weight (excluding the pneumatic control board). Second, the relatively slow response to pressure inputs make it inappropriate to implement the dynamic modeling of rigid parts. In contrast, prior research on soft pneumatic fingers has shown the feasibility of using geometric models for real-time position control~\cite{tawk2020position}. 

\subsection{Static Model}

\begin{figure*}[!h]
\vspace{6pt}
	\centering
	\includegraphics[width=0.8\linewidth]{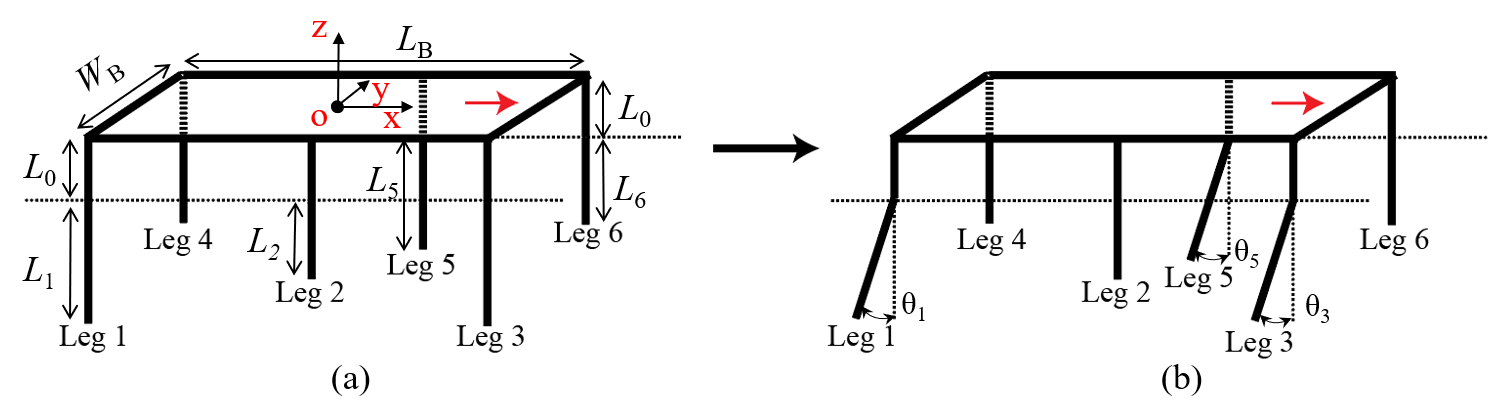}
	\vspace{-12pt}
	\caption{Modeling SoRX's quasi-static forward motion.}
	\label{fig:model}
	\vspace{-8pt}
\end{figure*}

In this work, we propose a static model based on geometric constraints for each leg (see Fig.~\ref{fig:actuator}b). We use one revolute and one prismatic joint to model the bending and extension parts, respectively. Parameters $\theta_i$ and $L_i$ are used to denote joints' values for leg $i=1,\ldots,6$. Note that $L_i$ include both the length of the extension part and the distance to the cut of the bending part. Using the model for single legs, we can further model the whole robot (see Fig.~\ref{fig:model}). Six legs are connected to a planar body frame of length $L_B$ and width $W_B$. The length of leg $i$ can be written as $L_0 + L_i$ where the $L_0$ is a constant that denotes the distance from the bending part to the robot's frame. The Euler angles of the robot planar frame are used to represent the robot's orientation. 

By design, there are two steady states for a single tripod gait: 1) only the extension part actuated (Fig.~\ref{fig:model}a), and 2) both parts actuated (Fig.~\ref{fig:model}b). In the first state, the extension parts of the tripod $\{1,3,5\}$ elongate and lift the body, then the bending parts are actuated and create angles $\{\theta_{1},\theta_{3},\theta_{5}\}$ to propel the robot forward. Both extension parts $L_i$ and bending parts $\theta_i$ depressurize when the other tripod actuates to support the robot.       

We compute the robot's height and orientation with respect to parameters $L_i$ and $\theta_i$. Note that we use the height of the geometric center of the robot's planar frame to denote the robot's height (point $o$ in Fig.~\ref{fig:model}a) as well as its Euler angles to represent the robot's orientation. Consider tripod $\{1,3,5\}$ is pressurized.  Then, the height of the robot can be written as  
\begin{equation}
\label{eq_height}
h=L_0 + \frac{L_1 + L_5}{2}\enspace.
\end{equation}
By design, we set $L_1 = L_3,\quad L_4 = L_6$ in all phases of the alternating tripod gait. The robot's roll angle along $x$ axis is

\begin{equation}
\label{eq_orient}
\phi = \text{atan}(L_5 - L_1, \frac{W_B}{2})\enspace.
\end{equation}



\subsection{Pressure Model and Parameter Identification}
A major contribution in the results we present in this work is that we implement a feedback pressure control for precise pneumatic regulation (to be elaborated in Section~\ref{control}). To derive that controller, it is first crucial to determine the relation between model parameters $L_i$ and $\theta_i$ with pressure $p$, which is needed for the robot's feedforward position control. 
Deriving analytically an accurate model of air dynamics in the actuators can be quite complicated; yet, examining the measured experimental data as a function of input air pressure, we can approximate the model using polynomials. 

To determine the relation between input pressure and output leg length, we perform a series of extension tests. We place the robot on flat ground, pressurize the extension part of the legs within a single tripod, and record the pressure\footnote{Gauge pressure sensors are used throughout this work. Both desired and measured pressure values are relative to atmospheric pressure.} (kPa) as well as the length (mm) of the actuated legs in steady state. 
Since the robot's legs are not massless and the length of the extension parts is sensitive to the load, preliminary testing revealed asymmetries to the response of the extension parts on the two sides of a tripod. 
To study this asymmetry within a tripod, we thus test the two sides of a tripod (i.e. the side with one leg and the other side with two legs) separately. 
Within these two cases, we further study two sub-cases in which the legs of the not-active side are either not actuated or pressurized at a constant pressure of 30 kPa, which is used in the experiments. 
The four considered cases and their respective notations are contained in Table~\ref{tab:model_cases}. Note that in double-leg cases, we measure the length of both legs and record the average.

\begin{table}[!th]
\vspace{0pt}
    \caption{Test Cases for Extension Part Modeling}
    \vspace{-12pt}
    \label{tab:model_cases}
    \begin{center}
    \renewcommand{\arraystretch}{1.5}
    \begin{tabularx}{\columnwidth}{l l}  
        \toprule
        one$_{w/o}$
        & Single-leg tripod side actuated, other side not actuated \\
        one$_{w/}$
        & Single-leg tripod side actuated, other side pressurized ($30$\;kPa) \\ 
        two$_{w/o}$
        & Double-leg tripod side actuated, other side not actuated \\ 
        two$_{w/}$
        & Double-leg tripod side actuated, other side pressurized ($30$\;kPa) \\
        \bottomrule
    \end{tabularx}
    \end{center}
    \vspace{-12pt}
\end{table}

\begin{figure}[!h]
\vspace{-7pt}
	\centering
	\includegraphics[width=0.9\linewidth]{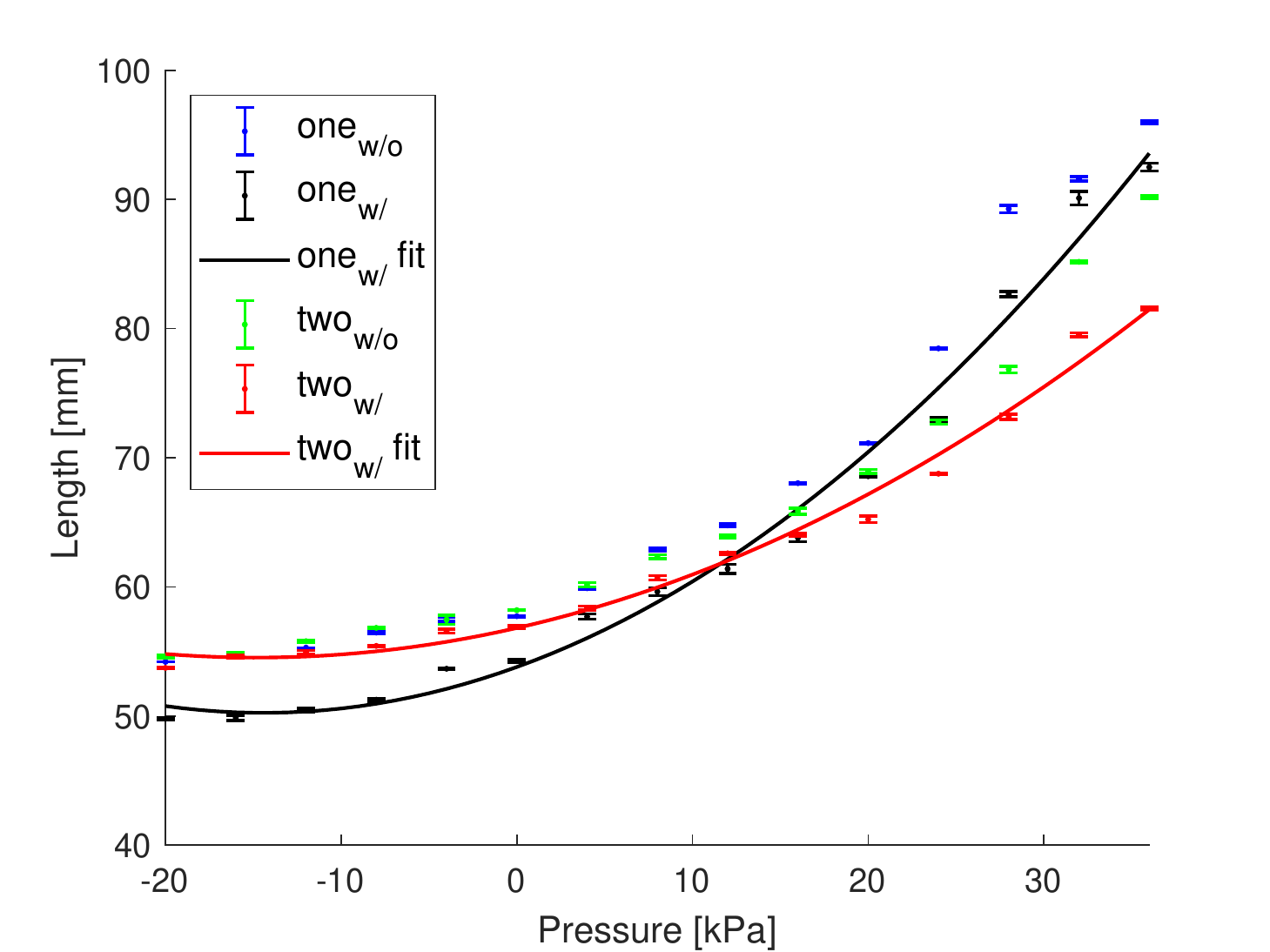}
	\caption{Relations between input pressure and output leg length.}
	\label{fig:fit}
	\vspace{-3pt}
\end{figure}

We apply pressure inputs ranging from $-20$\;kPa to $36$\;kPa with sampling interval of $4$\;kPa.\footnote{Actuators have higher risk to break when input pressure exceeds $36$\;kPa.}
Four distinct measurements are taken for every sampled pressure input. Figure~\ref{fig:fit} depicts mean values and one-standard deviations for all four cases shown in Table~\ref{tab:model_cases}. 
Experimental results confirm asymmetries on two sides of the tripod. Moreover, the double-leg tripod side is observed to have larger decrease in the elongation with the same positive pressure when the other side pressurized while the single-leg tripod side displays a larger decrease in length with the same negative pressure.    

We select to approximate relations where the other sides are actuated ($\text{one}_{\text{w}},\text{two}_{\text{w}}$) as the pressure models since two sides of the tripod are actuated for most of the tests. Experimental results show that the relations can be approximated by second-order polynomials. The curves are plotted in Fig.~\ref{fig:fit} as $\text{one}_{\text{w/}} \text{fit}$ and $\text{two}_{\text{w/}} \text{fit}$. On the other hand, for the angle $\theta$ model, we approximate the relation between input pressure and  bending angle $\theta$ presented in our previous work~\cite{liu2020sorx}. Polynomial coefficients for all models are listed in Table~\ref{table_coeff}. $R^2$ values of the three models are calculated to validate the fitting performance; $R^2_{\text{one}_{\text{w/}}} = 0.9877$, $R^2_{\text{two}_{\text{w/}}} = 0.9878$ and $R^2_{\theta} = 0.9691$. 

\begin{table}[!h]
\renewcommand{\arraystretch}{1.3}
\vspace{-6pt}
\caption{Polynomial Coefficients for Model Fitting}
\vspace{-6pt}
\label{table_coeff}
\centering
\begin{tabular}{c|c|c|c}
\hline
\bfseries Models & \bfseries Polynomials &\bfseries Units  &\bfseries Ranges\\
\hline\hline
$\text{one}_{\text{w/}}$ & $0.017p^2+0.492p+53.801$ & mm & [-20, 36] kPa\\
\hline
$\text{two}_{\text{w/}}$& $0.010p^2+0.309p+56.821$ & mm  &[-20, 36] kPa\\
\hline
$\theta$ & $0.010p +0.0153$ & rad &[-20, 50] kPa\\
\end{tabular}
\end{table}

\section{Controller Design} \label{control}

\subsection{Pneumatic Regulation Board}
In our prior work~\cite{liu2020sorx}, SoRX was driven by a modified version of an open-source pneumatic control board~\cite{holland2014soft}. In that board, every air output channel was connected to two pairs of valves and pumps to allow for both pressurization and depressurization. Instead of free-flow passive deflation, active depressurization significantly improves the walking performance since it can accelerate bending legs to recover to upright configurations. At the same time, active depressurization can further shorten the extension parts, thus increasing foot clearance to facilitate overcoming obstacles. 
The pneumatic regulation board proposed herein builds upon principles of the previous configuration and also includes pressure sensors to provide feedback. 


\begin{figure}[!h]
	\centering
	\includegraphics[width=0.9\linewidth]{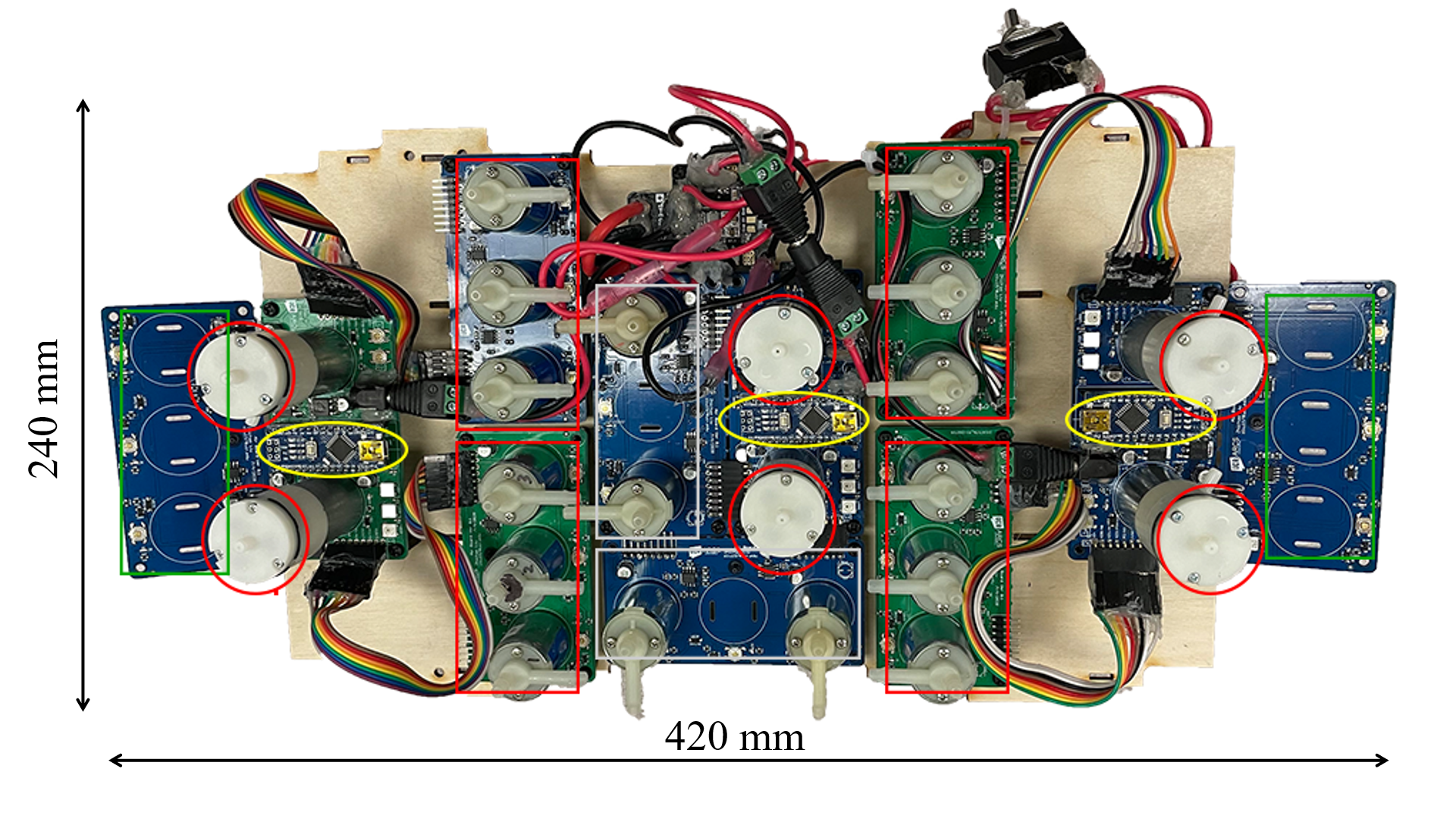}
	\caption{Top view of the developed pneumatic regulation board. Core components are highlighted and discussed in the text.}
	\label{fig:board}
	\vspace{-12pt}
\end{figure} 

In this work, we use custom printed circuit boards (PCBs) for the pneumatic regulation board to minimize size and weight. The PCB design is based on a portable open-source pneumatic controller\,\footnote{https://github.com/Programmable-Air} with minor changes to the operational amplifier circuit for pressure sensors. A top view of our developed pneumatic regulation board is shown in  Fig.~\ref{fig:board}. There are in total six pumps (red circles) and 16 solenoid valves on the board. Half of them are used for pressurization; the other half are responsible for depressurization. There are three types of valve boards: 1) boards with three valves and one pressure sensor (red box), 2) boards with two valves and one pressure sensor (white box), and 3) boards with only one pressure sensor (green box). Three micro-controllers (Arduino Nano, yellow ellipses) coordinate with the companion computer (Odroid XU4 [not shown in image]) to read pressure values as well as control valves and pumps. Electronics are powered by a 3500 mAh 3-cell LiPo battery. The board has a compact design ($240$\;mm\,L\,x\,$420$\;mm\,W\,x\,$140$\;mm\,H), and weighs $1.7$\;kg. The board is about half the size of the one used in~\cite{liu2020sorx} but with twice the number of output channels. The board is fitted with casters for portability and ease of use in experiments (Fig.~\ref{fig:test}). 

\begin{figure}[!h]
	\centering
	\includegraphics[width=0.7\linewidth]{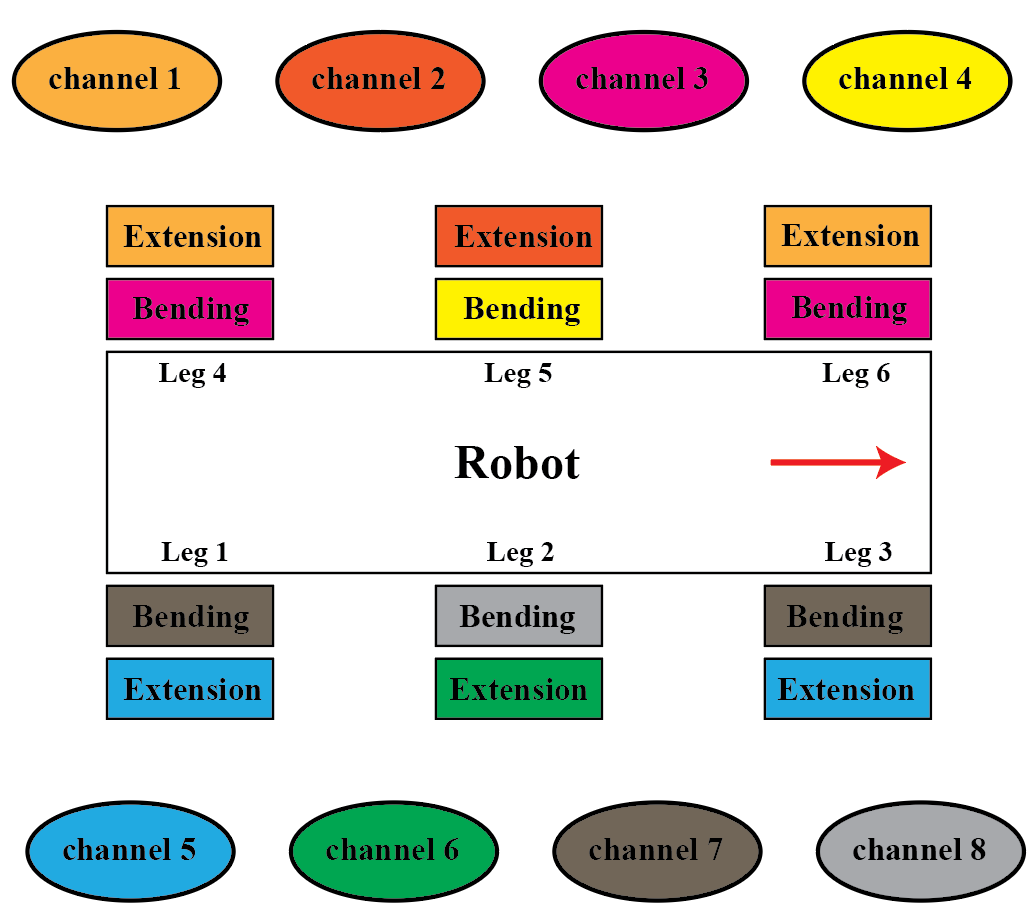}
	\caption{A diagram of eight air output channels to drive 12 actuation parts (six extension parts and six bending parts). Actuation parts and air output channels in the same color are connected, i.e channel 1 is connected to extension parts of leg 4 and leg 6. (Figure best viewed in color.)}
	\label{fig:tube}
	\vspace{-12pt}
\end{figure}

Compared to the only four air output channels that were actuated in our previous work~\cite{liu2020sorx}, the pneumatic regulation board in this paper implements eight channels in total to introduce more motion capabilities for SoRX (specifically, body orientation and turning). Four additional channels are used to address the body orientation control and turning (to be elaborated in Section~\ref{turn}). 
Figure~\ref{fig:tube} depicts the air flow logic for the pneumatic regulation in this work. There are in total eight air output channels (shown in different colors), and six legs each comprising two actuated parts (extension and bending parts). The channels and actuation parts of same color are connected. By design, the extension parts and the bending parts of the two outer legs on the same side are connected\,\footnote{Note that the two parts of the same leg are not connected.} and operated with the same pressure input (that is, pairs \{Leg 1 \& Leg 3\} and \{Leg 4 \& Leg 6\}). The two parts of the middle legs (i.e. Leg 2 and Leg 5) are separately actuated with four additional channels. 
 
\subsection{Pressure Feedback Controller}
In our pneumatic regulation board,  pressurization and depressurization are attained by different pairs of pumps and valves. Because of this, there can be significant delays when transitioning between actuation modes. Existing feedback control methods (e.g., PID controllers) based on pressure values alone failed in our preliminary experimental tests, causing oscillations when the pressure is close to zero.

To mitigate this challenge, we propose herein a feedback controller to achieve relatively fast and precise pressure control and avoid oscillations.  In our design, desired trajectories of each air output channel consist of two values: \texttt{mode} and \texttt{desired}. 
We command the \texttt{mode} to be either \texttt{pressurize} or \texttt{depressurize}, and the \texttt{desired} to be desired pressure values in the steady state. 

\begin{algorithm}[!h]
\label{alg}
\SetAlgoLined
\textbf{Given:} total time $T$,  threshold $\epsilon$\;
\textbf{Input:} the desired trajectories (mode$_t$, desired$_t$), pressure feedback values real$_t$ at time $t$\;
 \While{$t\le T$}{
  \eIf{mode$_t$ == pressurize}{
   \If{real$_t < $ desired$_t$ - $\epsilon$  and  Valve$_1$ == closed}{
   Valve$_1\leftarrow$ open\;  
   Pump$_1\leftarrow$ on\;
   }
   
   \If{real$_t \ge  $ desired$_t$ and Valve$_1$ == on}{
   Valve$_1\leftarrow$ closed\;
   Pump$_1\leftarrow$ off\;
   }
   }{
   \If{real$_t > $ desired$_t$ + $\epsilon$  and  Valve$_2$ == closed}{
   Valve$_2\leftarrow$ open\;
   Pump$_2\leftarrow$ on\;
   }
   
   \If{real$_t \le  $ desired$_t$ and  Valve$_2$ == on}{
   Valve$_2\leftarrow$ closed\;
   Pump$_2\leftarrow$ off\;
   }
  }
 }
 \caption{Pressure Feedback Controller}
\end{algorithm}

The pseudo code for single-channel pressure feedback control is detailed in Alg.~\ref{alg}. Two pumps and two valves  contribute to the regulation of each air output channel. Let Valve$_1$ and Pump$_1$ be used for pressurization while the rest take charge during depressurization. All pumps and valves are closed by default. Note that the algorithm uses a threshold $\epsilon$ to avoid oscillations. Thresholds for each channels are empirically tuned. In general, the bending parts are more sensitive to pressure changes, therefore larger thresholds are applied therein. 

\begin{figure}[!h]
\vspace{-12pt}
	\centering
	\includegraphics[width=0.75\linewidth]{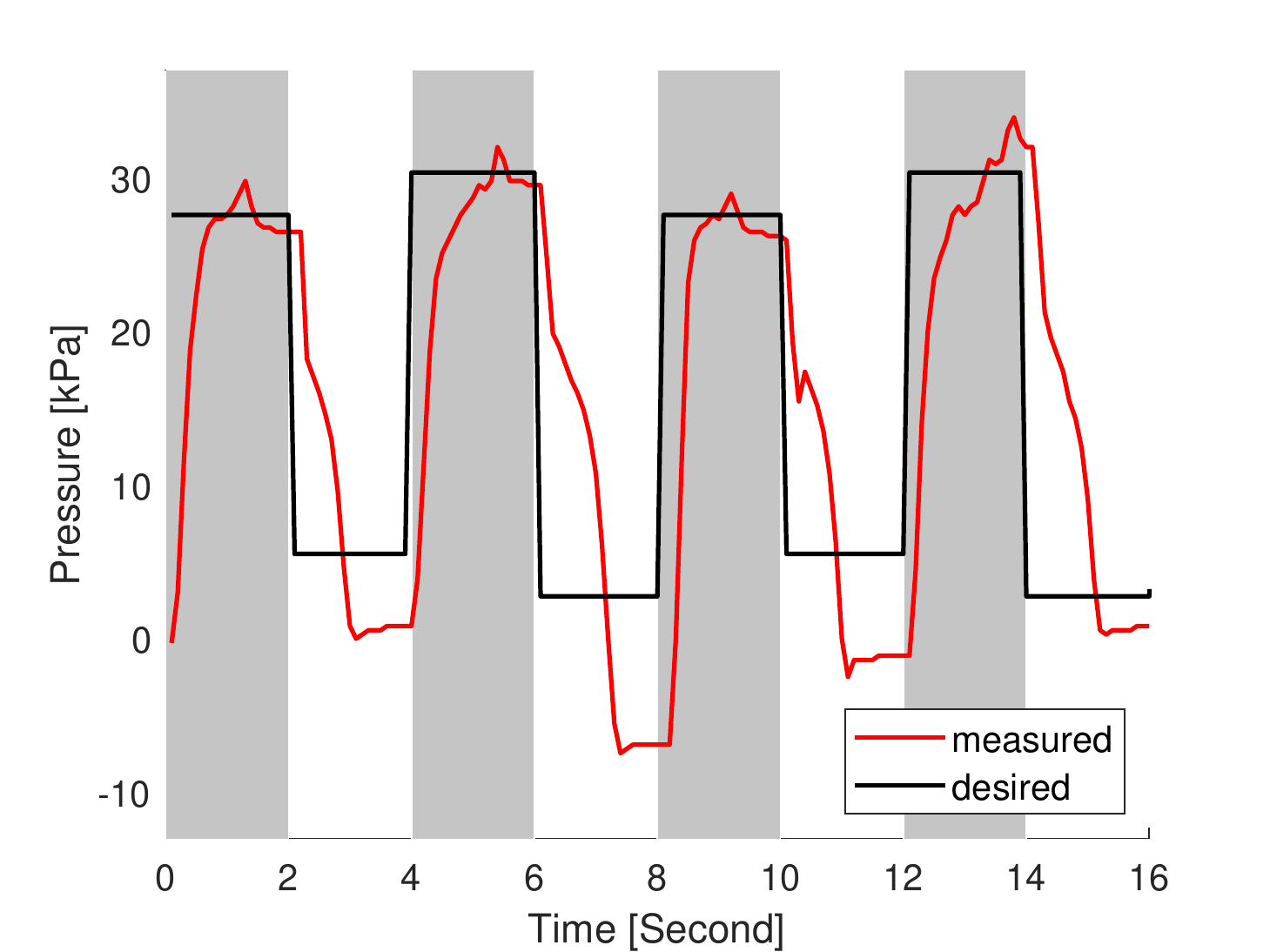}
	\vspace{-6pt}
	\caption{Step response for the proposed pressure feedback controller. The desired air pressure values are plotted in black while measured ones in red. The grey boxes represent that the \texttt{mode} is \texttt{pressurize} while the white ones denote \texttt{depressurize}.}
	\label{fig:response}
	\vspace{-9pt}
\end{figure}

The performance of the pressure feedback controller is evaluated by a step response test. In the experiment, a single extension part was actuated to track step trajectories with the proposed pressure feedback controller. The desired and measured air pressure values are shown in Fig.~\ref{fig:response}. The grey boxes represent that the \texttt{mode} is \texttt{pressurize} while the white ones denote \texttt{depressurize}. 

From the figure, the measured pressure in the steady state is generally tracking the positive desired one with small overshoot. However, when the desired pressure is close to or smaller than zero, large tracking errors are observed in the steady state. Mismatches in negative pressure are caused because pressure decreases very fast when the volume of the air chamber is close to its minimum. However, based on Fig.~\ref{fig:fit}, negative pressure values have little impact to the overall leg length. Hence, we consider that tracking errors, when the desired pressure is close to or less than zero, have acceptable impact to the motion control of the robot.         

\section{Trajectory Tracking}\label{turn}
\subsection{Walking and Turning}
\begin{figure}[!h]
\vspace{1pt}
	\centering
	\includegraphics[width=0.9\linewidth]{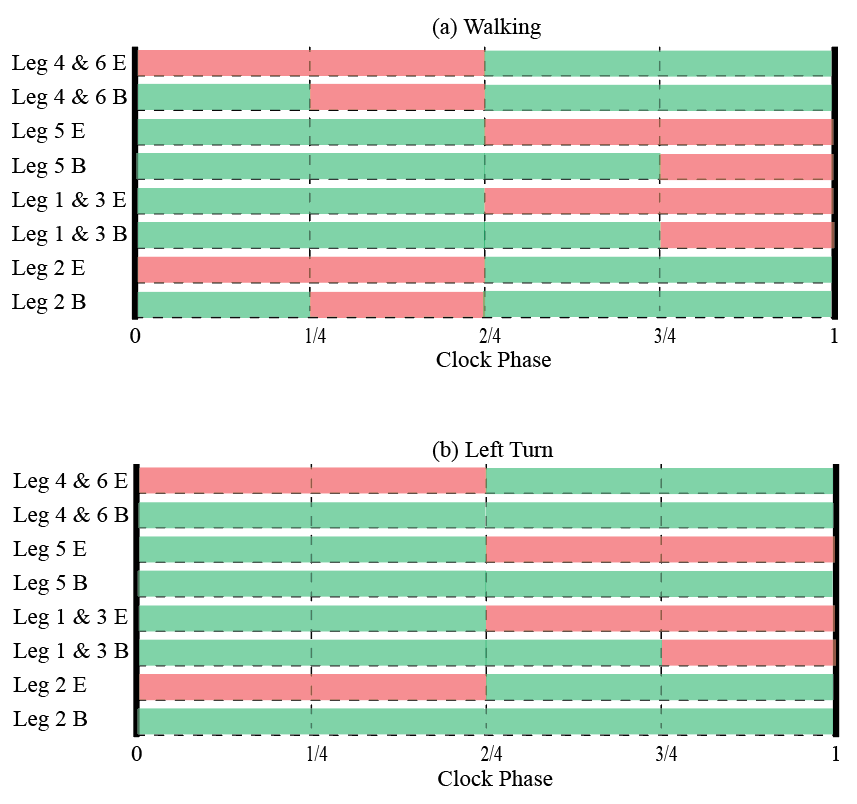}
	\vspace{-9pt}
	\caption{Actuation sequences for (a) walking and (b) left turn. (Figure best viewed in color.)}
	\label{fig:turn}
	\vspace{-15pt}
\end{figure}

This paper adopts the same actuation sequence as in~\cite{liu2020sorx} for walking (see Fig.~\ref{fig:turn}a). Notations \texttt{E} and \texttt{B} represent extension and bending parts, respectively. Red boxes are used to represent pressurization, while the green ones stand for depressurization. In the walking task, each tripod is actuated for half of the clock phase. During the actuation of each tripod, the extension parts are pressurized first and hold the pressure, followed by pressurization of the bending parts.  

Compared to rigid robots, SoRX relies on leg's shape morphing to move, thus existing turning methods for hexapedal and octapedal robots with coupled leg motion (e.g.,~\cite{saranli2001rhex,karydis2014planning,karydis2016navigation}), were not successful in our preliminary experimental tests. To this end, we adopt in this work a simple yet effective turning method for the robot. Figure~\ref{fig:turn}b shows a sample actuation sequence for making a left turn. Actuation sequences for the extension parts remain the same as in normal walking, however, only the bending parts of two legs opposite to the turning direction are actuated. The difference in the actuation of bending parts within a tripod enables the robot to turn while the elongation of the extension parts of the other tripod assists legs to recover to upright configurations.    

\begin{figure}[!h]
\vspace{-7pt}
	\centering
	\includegraphics[width=0.75\linewidth]{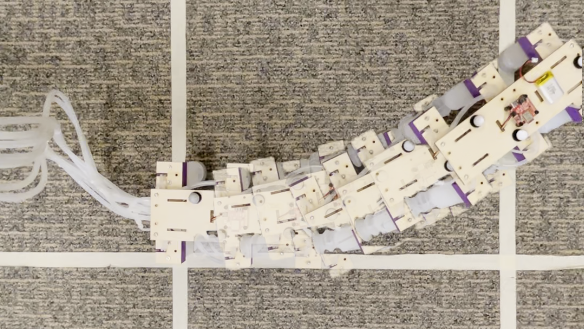}
	\vspace{-6pt}
	\caption{Composite images of a sample test on turning. }
	\label{fig:turn_traj}
	\vspace{-12pt}
\end{figure}

We test the performance of the proposed turning method with consecutive left turns. Figure~\ref{fig:turn_traj} shows snapshots from a sample turning trajectory of the robot. Observations suggest that the bending part of leg 5 still curves passively due to the weight, however, actuation of two legs on the other side enables the robot to turn. A full actuation sequence enables the robot to turn by approximately $10^\circ$.  

The robot's walking speed is determined via the time of a clock phase in Fig.~\ref{fig:turn}. To achieve accurate pressure control, a longer phase ($6.6$\;sec) is used compared to the one in~\cite{liu2020sorx} ($1.6$\;sec). As a result, the pressure feedback-enabled walking speed of the robot (without turnings) is approximately $24.5$\;mm/s ($0.11$\;BL/s), compared to the open loop speed of $101$\;mm/s ($0.44$\;BL/s) reported in \cite{liu2020sorx}. When turning, the speed of the robot is further slowed down due to the fact that the robot moves forward during only half of the clock phase.

\subsection{Closed-loop Trajectory Tracking}
\begin{figure}[!h]
\vspace{-6pt}
	\centering
	\includegraphics[width=0.7\linewidth]{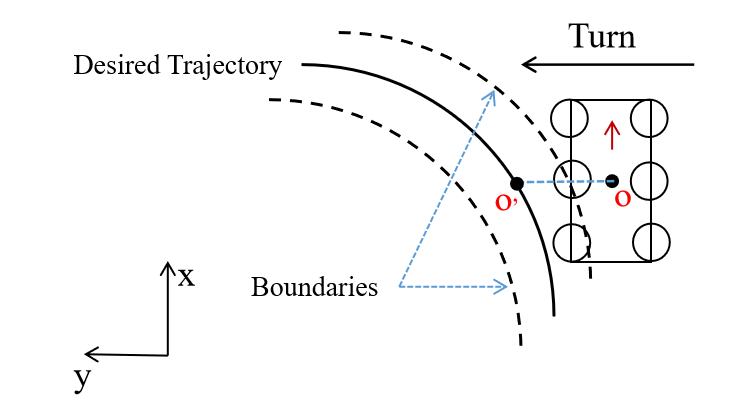}
	\caption{Illustration of the closed-loop trajectory tracker.}
	\label{fig:tracking}
	\vspace{-4pt}
\end{figure}

The significance of the developed turning method is that it enables implementation of \emph{closed-loop trajectory tracking control} for the first time in the context of soft legged robots like SoRX. The approach we present herein is a direct and effective means that relies on trajectory corridors; investigation of tracking more involved trajectories in obstacle-cluttered environments is part of future work. Consider a desired trajectory containing 3D positions ($x,y,z$) as shown in Fig.~\ref{fig:tracking}. Along with the desired trajectory, we prescribe a 2D corridor (black dashed lines), which is defined to lie at a fixed, user-defined distance from the desired trajectory's projection on the x-y plane. 


The robot receives location data from motion capture at 100 Hz and compares the 2D position (the geometric center of the planar body) with the boundaries of the corridor at a rate of 10 Hz. When the center is located outside of the boundaries, the robot will trigger the turning method to move toward the desired trajectory, until the center is found across the desired trajectory. For instance, Fig.~\ref{fig:tracking} is sketched to show the center (point $o$) being outside the right boundary, thus the turning method drives the robot to turn left. 
Given the current location (point $o$) from the motion capture system, we map it to the desired trajectory (point $o^,$). In this work, point $o^,$ is found using the same $y$ values for simplicity as the robot tracks a straight line alone $x$ axis, however, the minimal distance with coordinate transformations can be used for mapping complex trajectories in the future work. The height of the mapped point $z(o^,)$ is used as the desired height of the robot at the current location. The desired air pressure is calculated based on models in Table~\ref{table_coeff}, and sent to the pressure feedback controller.    

\begin{table}[!h]
\vspace{-6pt}
\renewcommand{\arraystretch}{1.3}
\caption{Key Parameters and their Values}
\label{table_value}
\vspace{-3pt}
\centering
\begin{tabular}{c c c c c}
\toprule
 $L_B$ & $W_B$ & $L_0$ & $\epsilon_B$& $\epsilon_E$\\
\midrule
$230$\; mm & $140$\; mm & $65$\; mm & $10$\; kPa  &  $5$\; kPa\\
\bottomrule
\end{tabular}
\vspace{-6pt}
\end{table}

\section{Experimental Results}
We conduct both indoor and (proof-of-concept) outdoor experiments. In indoor tests, we evaluate the proposed model-based position control and closed-loop trajectory tracking on the SoRX robot~\cite{liu2020sorx}. The position of the robot is captured using a 12-camera Optitrack motion capture system. A desktop (Intel NUC 10 with 2.3 GHZ i7 CPU) is used as the companion computer. The robot operates on flat ground. Values for key parameters used in the paper are listed in Table~\ref{table_value}. Note that $\epsilon_B$ and $\epsilon_E$ are the thresholds for bending and extension parts used in Alg.~\ref{alg}, respectively. In outdoor tests, we evaluate the preliminary feasibility of manually controlled navigation over unstructured terrain for the robot.

\subsection{Position Control}
Two experiments are conducted to evaluate the proposed static models in Section~\ref{sec_model}. In the first test, the robot is placed on the ground, and one tripod is controlled to change the height of the center (point $o$). The largest desired height of $132$\;mm is achieved when all extensions parts are pressurized while the lowest desired height of $120$\;mm corresponds to the state of depressurization of the tripod.

Desired pressure values are determined based on~\eqref{eq_height} and the polynomials models in Table~\ref{table_coeff}. It is important to note that legs for both two sides of the tripod have the same length by design. Based on the fitting models, we calculate the max and min pressure values $19.75$ and $-8.11$ \;kPa for the extension parts on double-leg tripod side, while $16.93$ and $2.26$\;kPa for the single-leg side. We input the desired pressure values to the pressure feedback controller with a time interval of $2$\;sec and record the height from the motion capture system.      

\begin{figure}[!h]
\vspace{-9pt}
	\centering
	\includegraphics[width=0.7\linewidth]{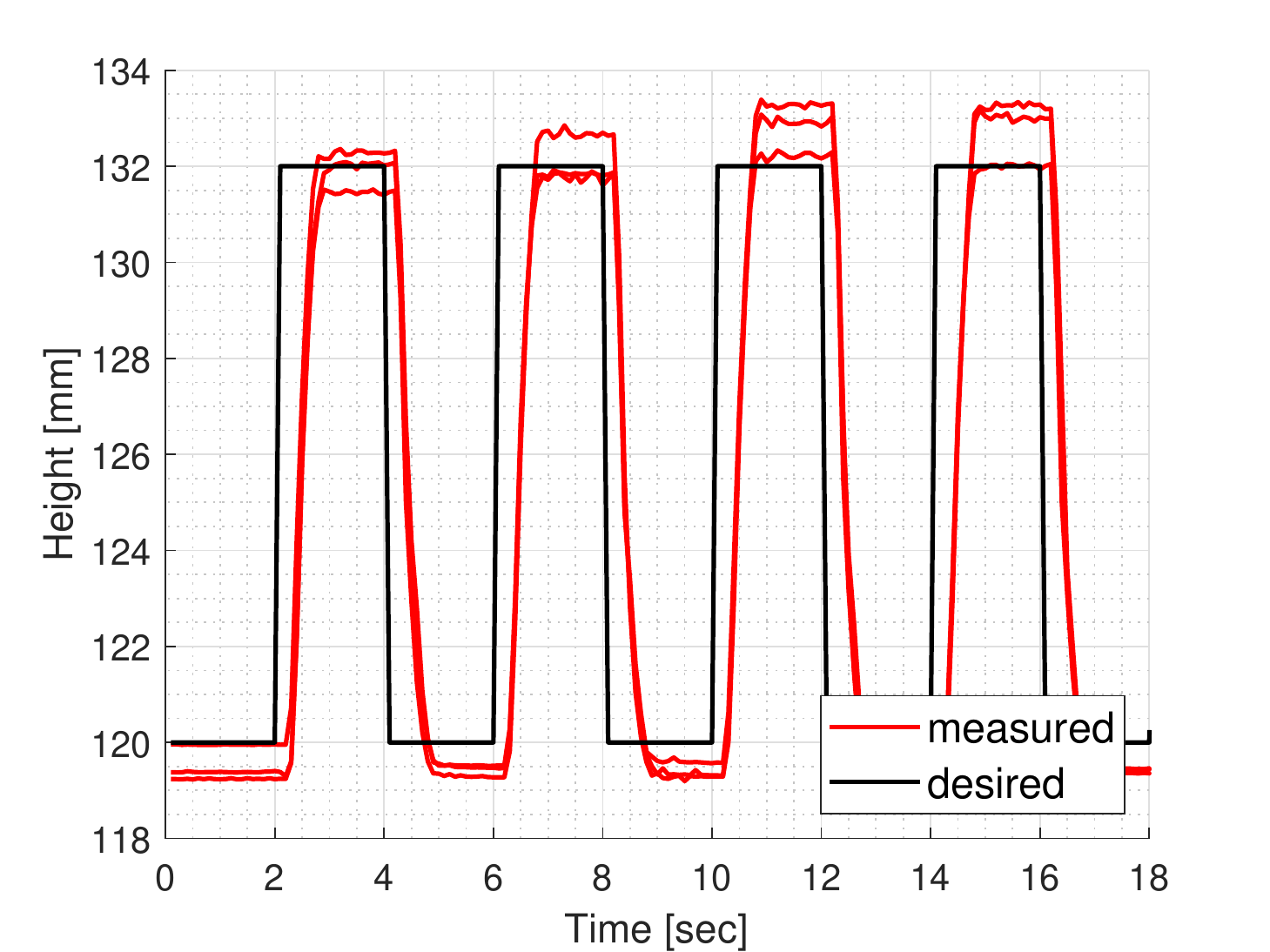}
	\vspace{-9pt}
	\caption{Desired and measured values for the height of the robot's center (point $o$). }
	\label{fig:height}
	\vspace{-6pt}
\end{figure}

Figure~\ref{fig:height} presents both desired and measured height of the robot's center for three consecutive experimental trials. Although delays and relatively small steady errors are observed, results suggest the height of the center is tracking the desired trajectories with the proposed methods.

\begin{figure}[!h]
	\centering
	\includegraphics[width=0.7\linewidth]{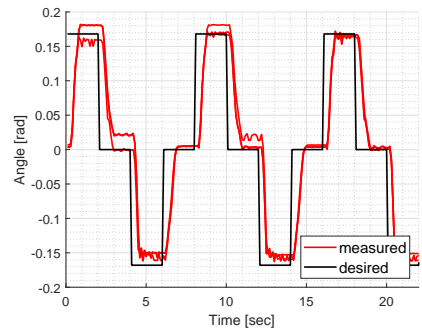}
	\caption{Desired and measured the roll angle $\phi$ for the robot's planar body.}
	\label{fig:roll}
	\vspace{-10pt}
\end{figure}

Similarly, we apply the same desired pressure inputs to evaluate~\eqref{eq_orient}. Given the difference between two extreme heights ($L_5 - L_1 = 12$\; mm), we can calculate the roll angle $\phi = \arctan{(2(L_5 - L_1)/W_B)} = 0.17$\; rad. Three consecutive tests are conducted and results are shown in Fig.~\ref{fig:roll}. The measured roll angles are in general tracking the desired ones despite delays and steady errors introduced by the pressure controller and model fitting.      


\subsection{Trajectory Tracking}

We conduct two experiments to validate the proposed closed-loop trajectory tracking control. In the first test, only the 2D position of the robot is considered. We command the robot to track two planar trajectories: 1) a straight line, and 2) a quarter circle. 

\begin{figure}[!h]
\vspace{-12pt}
	\centering
	\includegraphics[trim={0cm 0cm 0cm 0.5cm},clip,width=0.9\linewidth]{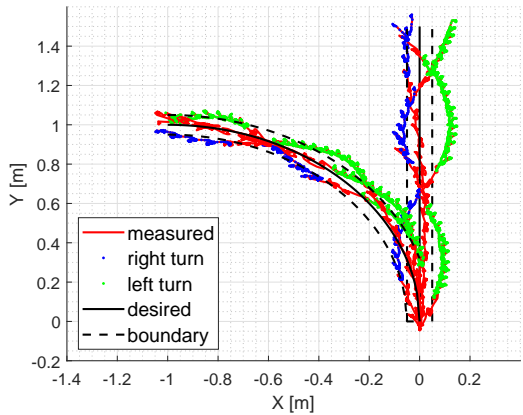}
	\vspace{-4pt}
	\caption{Results for closed-loop 2D trajectories tracking experiments.}
	\label{fig:track}
	\vspace{-6pt}
\end{figure}

In the straight line case, the robot starts at the origin and is expected to reach the point $(0, 1.5)$\;m; the robot stops after reaching the line $y = 1.5$\;m. The boundaries are set at $x = \pm 0.05$\;m. Three consecutive experimental trials are made with different starting angles (0, $\pm 15^\circ$). The desired and measured trajectories for all trials are shown in Fig.~\ref{fig:track}, where the blue and green dots denote components of the robot trajectory during which the tracker sends right and left turning commands, respectively. Results show that the robot walks generally in straight line without steering control with a zero starting angle, until reaching a distance of $1.2$\;m followed by right turns. Further, the effectiveness of the method is validated with $\pm 15^\circ$ starting angles. Results show that the robot walks outside of the boundaries shortly after the start, however, the trajectory tracking method drives the robot to move toward the desired trajectory with repeating changes of right/left turning sequences. 

A desired trajectory of quarter circle $(x+1)^2 + y^2 = 1, x\in [-1, 0]$ is set for the second experiment. Similarly, two boundaries $(x+1)^2 + y^2 = (1 \pm 0.05)^2$ are selected to trigger turning. The desired trajectory begins at the origin and moves toward the destination $(-1,1)$\;m, where the robot stops after reaching the line $x = -1$\; m. Three experimental trials are conducted with zero starting angles. Results in Fig.~\ref{fig:track} demonstrate that the proposed method enables SoRX to track both straight-line and turning trajectories. 

\begin{figure}[!h]
    \vspace{-9pt}
	\centering
	\includegraphics[trim={0cm 0cm 0cm 0.5cm},clip,width=0.75\linewidth]{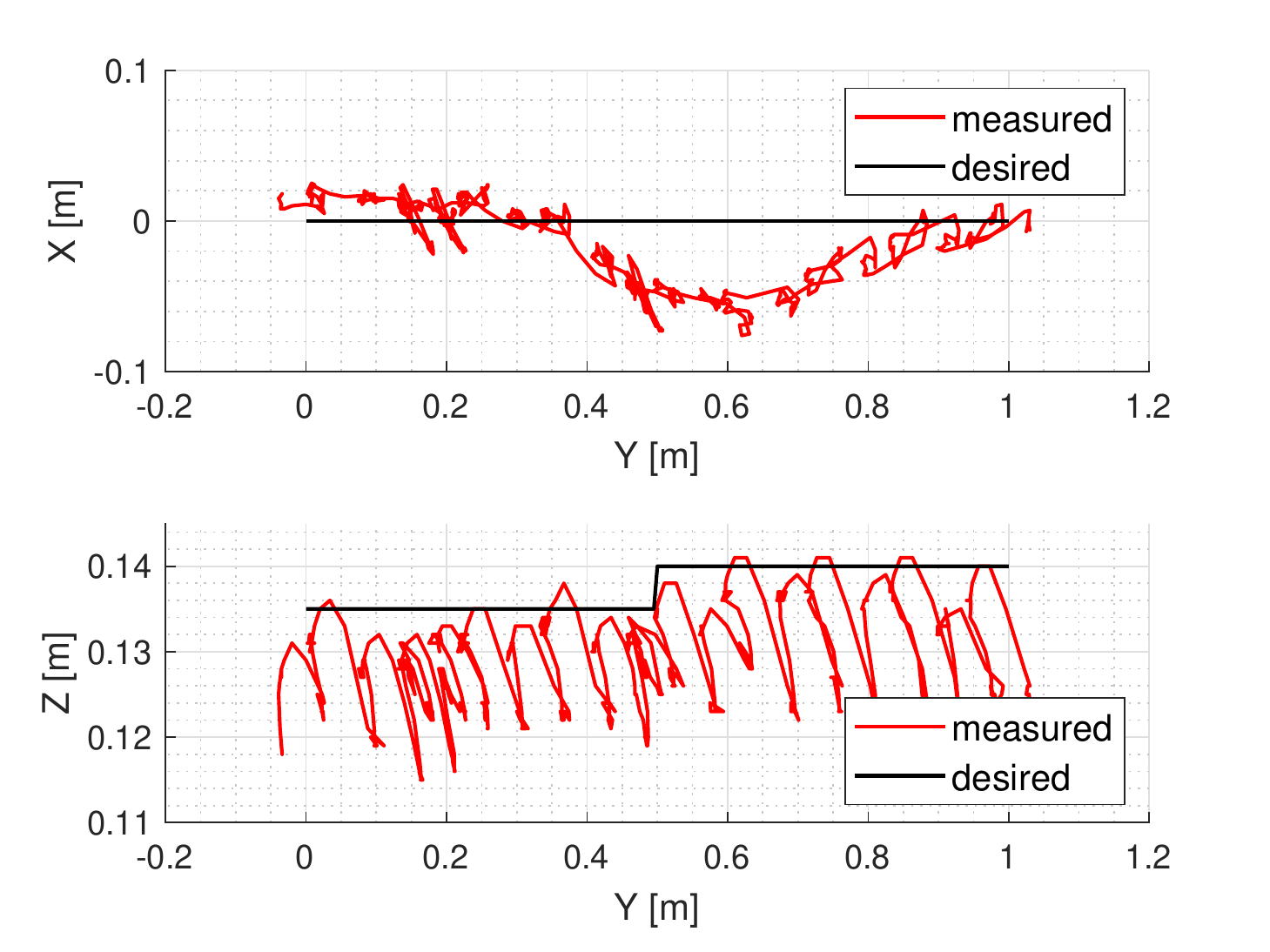}
	\vspace{-10pt}
	\caption{Desired/measured positions in variable-height trajectory tracking.}
	\label{fig:vary}
	\vspace{-12pt}
\end{figure}

For the second experiment, we command the robot to track a variable-height trajectory. The trajectory consists of a planar straight line from the origin to the point $(0,1)$\;m, with the desired maximal height switching from $0.135$\; to $0.140$\;m after reaching the line $y = 0.5$\;m. Tests are made with zero starting angles. The desired and measured trajectories of the robot are shown in Fig.~\ref{fig:vary}. Given the two steady states for the walking task, oscillations in the height of the robot's center are observed along the trajectories. However, results indicate the utility of our method to track variable-height trajectories since the maximal heights of the robot's center switch after passing the line $y=0.5$\;m, as desired. 

\subsection{Tracking Performance}
We list all tracking errors in Table~\ref{errors} for all experiments including pressure feedback control, position control, and trajectory tracking. Note that we use the distance of the measured positions to the desired trajectories for both line and curve tracking experiments. For instance, $d_1$ denotes the absolute value of the measured $x$ for the straight line tracking test. For the variable-height trajectory tracking test, we list the 2D straight line tracking error $d_3$, as well as the the height difference $h_2$ between desired and measured values for the locally maximal points.  

\begin{table}[!h]
\vspace{-4pt}
\renewcommand{\arraystretch}{1.3}
\caption{Tracking Performance}
\label{errors}
\centering
\begin{tabular}{l r c}
\toprule
\bfseries Variables & \bfseries Tracking Errors  &\bfseries Units \\
\toprule
 Step Response $p$ & $-0.737 \pm 11.198 $  & kPa\\
\midrule
 Position Control Height $h_1$ & $-0.263 \pm 4.910$ &mm\\
\midrule
 Position Control Angle $\phi$ & $0.006 \pm 0.073$  & rad\\
 \midrule
 Trajectory Tracking Line $d_1$ & $0.029 \pm 0.019$ & m\\
 \midrule
 Trajectory Tracking Curve $d_2$  & $0.045 \pm 0.020$ & m\\
 \midrule
 Trajectory Tracking Variable-Height $d_3$ & $0.024\pm 0.020$ & m\\
 \midrule
 Trajectory Tracking Variable-Height $h_2$ & $-1.474\pm 2.245$ & mm\\
 \bottomrule
\end{tabular}
\vspace{-10pt}
\end{table}

\subsection{Preliminary Feasibility for Outdoor Field Testing}
\begin{figure}[!h]
\vspace{6pt}
	\centering
	\includegraphics[trim={0cm 0cm 0cm 0.5cm},clip,width=0.75\linewidth]{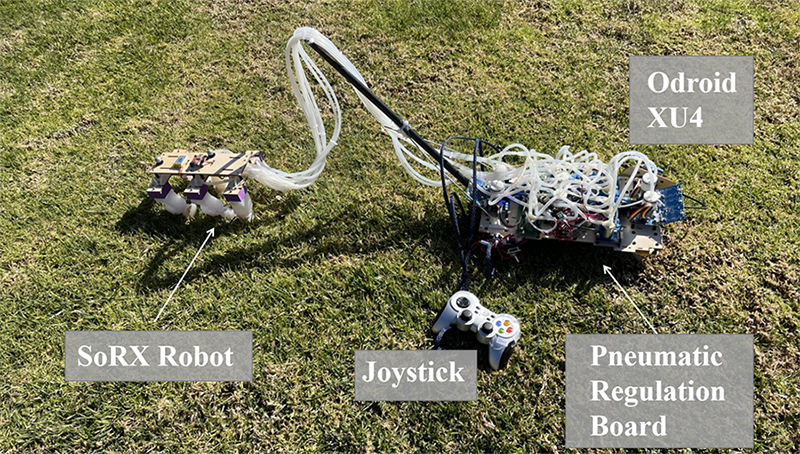}
	\vspace{-6pt}
	\caption{SoRX is able to operate outdoors while powered by the untethered pneumatic regulation board.} 
	\label{fig:test}
	\vspace{-18pt}
\end{figure}

Taking advantage of the compact and portable design of our pneumatic regulation board, SoRX can operate in outdoor environments. Figure~\ref{fig:test} shows a snapshot from preliminary field tests. An Odroid XU4 coordinates with the board; walking and steering is remote-controlled. Powered by the untethered board, SoRX operates on various types of natural rough terrain, including creeks and gravel (see Fig.~\ref{fig:cover}).    


\section{Conclusions}
This work contributes to extending the motion capabilities of a soft pneumatic legged robot SoRX, which has shown able to traverse rough, steep and unstable terrain~\cite{liu2020sorx}. Specifically, we proposed a static model based on geometric constraints for feedforward position control, and designed and implemented a compact and portable pneumatic regular board that powers up to eight channels of pressurization/depressurization with pressure feedback. We also introduced a pressure feedback controller, as well as a closed-loop variable-height trajectory tracking control method, that utilize the pneumatic regulation board to enable the robot to track straight-line and curving trajectories. Extensive experimental testing indoors revealed that the proposed methods can enable effective fully-pneumatic feedback trajectory tracking control for soft pneumatically-actuated legged robots. In addition, preliminary feasibility tests indicated that the developed board and controller can facilitate (remote-controlled) operation of the robot over unstructured terrain as well. We believe our work presents encouraging and repeatable results to motivate research on autonomous soft legged robots in outdoor environments. 

Future work includes implementation of a fully untethered soft legged robot, e.g., by integrating micro-fluidics-based circuits~\cite{hoang2021pneumatic}, and more involved trajectory trackers while considering interactions with the environment, and incorporation of sensors for fully-autonomous outdoor navigation. 

\section*{Acknowledgement}
The authors wish to thank Dr. Elia Scudiero for offering access to outdoor experimental fields for testing.



\bibliographystyle{IEEEtran}
\bibliography{IEEEabrv, ms}

\end{document}